%% file: paper.tex
\documentclass[runningheads]{llncs}
\usepackage{graphicx}

\usepackage[hang,flushmargin]{footmisc}
\usepackage[misc,geometry]{ifsym}
\usepackage{cite} 
\usepackage[T1]{fontenc}
\usepackage{amsmath}
\usepackage{mathrsfs}
\usepackage{graphicx}  
\usepackage{epstopdf}
\usepackage{bbm}
\usepackage{float}
\usepackage{multirow}
\usepackage{booktabs}
\usepackage{mathrsfs}
\usepackage{amssymb}
\usepackage[colorlinks,linkcolor=red]{hyperref}
\usepackage{siunitx}
\usepackage{tabularx}
\makeatletter

\newcommand{\Rmnum}[1]{\expandafter\@slowromancap\romannumeral #1@}
\makeatother
\usepackage{booktabs}
\usepackage{amsfonts,amssymb}
\usepackage{amsmath,bm}
\usepackage{array}
\newcommand{\PreserveBackslash}[1]{\let\temp=\\#1\let\\=\temp}
\newcolumntype{C}[1]{>{\PreserveBackslash\centering}p{#1}}
\newcolumntype{R}[1]{>{\PreserveBackslash\raggedleft}p{#1}}
\newcolumntype{L}[1]{>{\PreserveBackslash\raggedright}p{#1}}
\usepackage{algorithmicx,algorithm}
\makeatletter

\usepackage{xcolor}

\usepackage{comment}

\begin{document}





\title{SimEndoGS: Efficient Data-driven Scene Simulation using Robotic Surgery Videos\\ via  Physics-embedded 3D Gaussians}

\titlerunning{SimEndoGS}


\author{
Zhenya Yang,
Kai Chen,
Yonghao Long,
\and Qi Dou\textsuperscript{(\Letter)}
}


\authorrunning{Z. Yang et al.}

%

\institute{
    The Chinese University of Hong Kong, Hong Kong SAR, China\\ 
    \email{qidou@cuhk.edu.hk}
}

\maketitle

\input{sections/abstract.tex}
\input{sections/intro.tex}
\input{sections/method.tex}

\input{sections/experiments}

\section{Conclusion}
This paper introduces a novel framework based on Gaussian Splatting for automated surgical scene reconstruction from stereo surgical videos and physically-based endoscopic scene simulation with user-defined interactions. 
We coherently utilize the 3D Gaussian representation for reconstruction and simulation for convenient simulation, efficient visualization and realistic visual results.
We have specifically designed an optimization strategy to enhance the suitability of our learned 3D Gaussians for subsequent simulation tasks. 
Our method achieves a high degree of automation and superior efficiency, seamlessly transforming surgical videos into interactive simulation scenes. 
We anticipate that the integration of generative AI and 3D reconstruction techniques will inspire the development of interactive and highly realistic surgical scene generation, benefiting surgical training and enhancing the learning capabilities of surgical robots.


\bibliographystyle{splncs04}
\bibliography{reference}

\end{document}

%% file: sections/abstract.tex
\begin{abstract}
Surgical scene simulation plays a crucial role in surgical education and simulator-based robot learning. Traditional approaches for creating these environments with surgical scene involve a labor-intensive process where designers hand-craft tissues models with textures and geometries for soft body simulations. This manual approach is not only time-consuming but also limited in the scalability and realism. In contrast, data-driven simulation offers a compelling alternative. 
It has the potential to automatically reconstruct 3D surgical scenes from real-world surgical video data, followed by the application of soft body physics. 
This area, however, is relatively uncharted. In our research, we introduce 3D Gaussian as a learnable representation for surgical scene, which is learned from stereo endoscopic video. To prevent over-fitting and ensure the geometrical correctness of these scenes, we incorporate depth supervision and anisotropy regularization into the Gaussian learning process. Furthermore, we apply the Material Point Method, which is integrated with physical properties, to the 3D Gaussians to achieve realistic scene deformations.
Our method was evaluated on our collected in-house and public surgical videos datasets. Results show that it can reconstruct and simulate surgical scenes from endoscopic videos efficiently—taking only a few minutes to reconstruct the surgical scene-and produce both visually and physically plausible deformations at a speed approaching real-time. The results demonstrate great potential of our proposed method to enhance the efficiency and variety of simulations available for surgical education and robot learning.
\keywords{Soft Tissue Simulation \and Gaussian Splatting \and Surgical Video}

\end{abstract}

%% file: sections/intro.tex
\section{Introduction}
\label{sec:intro}
Endoscopic scene simulation is fundamental for surgical training, education and learning-based surgical robot automation~\cite{huang2023demonstrationguided,huang2023valueinformed,long2023humanintheloop}. 
Despite much efforts~\cite{hirota2003improved,bar2006simbionix,sofa} to simulate deformable objects in anatomical scenes, current solutions rely on manually designed textures that are time-consuming and not scalable. 
These textures often fail to capture the realistic appearance of various tissues and endoscopic illuminations in real data. 
Recent advancements in generative AI and 3D reconstruction techniques~\cite{rombach2022high,liu2024sora,hong2023lrm} have raised interest in developing an efficient data-driven surgical scene simulation pipeline, i.e., \emph{can we automatically generate photo-realistic and interactive scenes by only using surgical video}. 
One feasible way could be to first reconstruct 3D representation of a surgical scene from stereo video, then integrate physics into this scene representation for physically-based simulation.
However, achieving this goal is challenging.
\hspace{-1em}

Dynamic surgical scene reconstruction has been studied in recent works such as EndoNeRF~\cite{wang2022neural}, EndoSurf\cite{zha2023endosurf}, LerPlane\cite{yang2023neural} and ForPlane\cite{yang2023efficient}. 
All these NeRF-based methods typically requires sophisticated post-processing~\cite{tang2023delicate,wei2023neumanifold} to turn implicit representation to simulatable mesh~\cite{marching_cubes,molino2003tetrahedral}.
This transformation increases the time cost and lose the reconstruction detail of surgical scenes, making NeRF-based methods suboptimal for simulation in practice.
Very recently, Gaussian Splatting (GS)~\cite{kerbl20233d} has emerged as a promising alternative to NeRF\cite{mildenhall2021nerf}, offering superior 3D reconstruction results and faster inference speed overall.
Several concurrent works~\cite{zhu2024endogs,liu2024endogaussian,huang2024endo4dgs} have applied 3D-GS for surgical reconstruction. 
Unlike NeRF, 3D-GS uses explicit Gaussian representation, making it feasible for physical simulation without any modality transformation\cite{xie2023physgaussian}.
It should be emphasized that all dynamic surgical scene reconstruction methods mentioned above model the tissue deformation using neural networks which take time and spatial location as input. 
Due to the lack of consideration for material properties and physical rules, all these learning-based methods could not support interactive surgical simulation.


Meanwhile, simulating soft-body has been studied via various methods \cite{in2fem,SULSKY_mpm,muller2007position}.
Among them, Material Point Method (MPM)\cite{SULSKY_mpm} is well-suited for endoscopic scene simulation due to its realism and efficiency. 
MPM is a physically based simulation method that combines particles and grids to accurately model the deformable objects.
A prior work PhysGaussian~\cite{xie2023physgaussian} has promisingly shown the feasibility to simulate 3D Gaussians as deformable objects integrating MPM.
The well-defined nature of material point makes MPM suitable to simulate 3D Gaussian while the learned appearance features of 3D Gaussian enhance the visual realism of MPM.
Due to the compatibility between MPM and 3D Gaussian, the physics could be naturally embedded into 3D Gaussians by updating them using MPM.
Inspired by PhysGaussian, we hope to represent the endoscopic scene with 3D Gaussian and perform simulation on this representation.
However, due to the limited movement range of camera in narrow endoscopic space, the 3D Gaussians trained from surgical videos are prone to overfitting which may introduce artifacts in simulation.
Overcoming this problem is important for performing surgical scene simulation in a fully automated pipeline.


In this paper, we endeavor to simulate reconstructed surgical scenes
captured from single-viewpoint stereo endoscope in a completely data-driven manner (called SimEndoGS). 
To the best of our knowledge, this is the first work to utilize the emerging 3D Gaussian Splatting and Material Point Method (MPM) framework to address the challenge of surgical scene simulation.
We summarize our contributions as follows: 
\textbf{1)} An efficient and automated pipeline consisting of designed 3D-GS reconstruction module and a subsequent efficient MPM simulation module.
\textbf{2)} A geometric regularization is proposed to overcome overfitting issue of Gaussian Splatting on endoscopic data.
\textbf{3)} The Neo-Hookean model and an adapted MPM solver for 3D Gaussian are seamlessly integrated into our pipeline to generate physically-based deformation in surgical scenes. 
Experimental results on our robotic surgery videos demonstrate the success of our data-driven reconstruction and simulation pipeline, which can support soft-tissue interactions with realistic deformation and approaching real-time speed.



%% file: sections/method.tex
\section{Method}

\begin{figure}[t]
    \includegraphics[width=\textwidth]{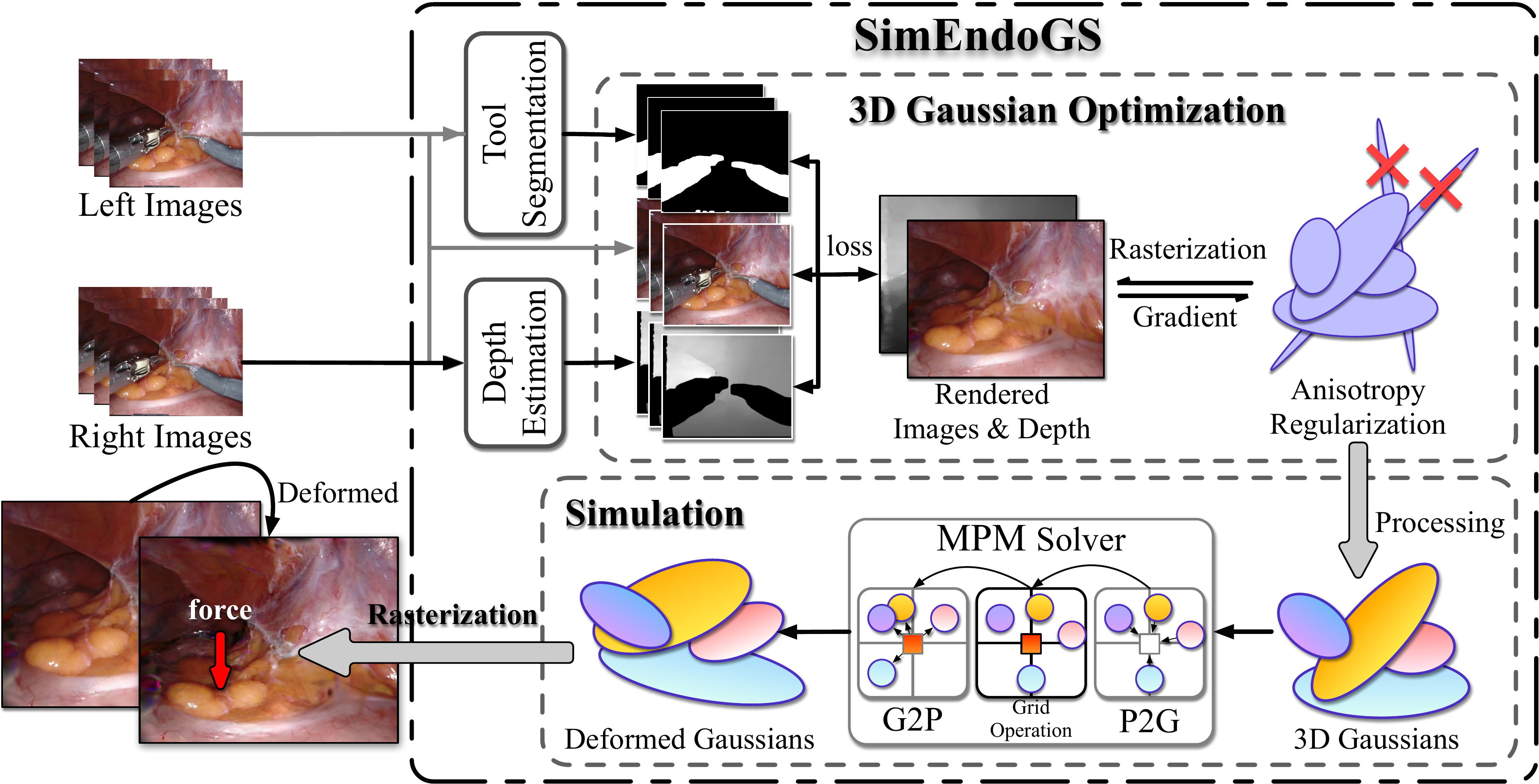}
    \caption{
        {\bf An overview of the proposed data-driven surgical simulation framework.}
        It consists of automatic scene reconstruction and physically-based scene simulation using 3D Gaussians.
        }
    \label{overview}
\end{figure}

\subsection{Overview of the Data-driven Scene Simulation Framework}
Let $\mathcal{V}=\{(\boldsymbol{I}^{l}_{i},\boldsymbol{I}^{r}_{i})\}^{T}_{i=1}$ be a stereo surgical video with $T$ frames, we aim to develop a fully automatic framework to reconstruct the simulation environment from the video and perform physically-based endoscopic tissue simulation with high fidelity.
First of all, we resort to the recent Gaussian Splatting technique for efficient surgical scene reconstruction in Section \ref{sec representation}.
It represents the surgical scene with a group of 3D Gaussians.
Similar to \cite{wang2022neural,yang2023efficient,zhu2024endogs}, we leverage a segmentation model to localize tool regions for each video frame.
The tool mask is used for occlusion-free surgical scene reconstruction.
Conventional GS-based reconstruction is optimized by minimizing the image-level difference between the reconstructed scene and the original video frame.
When it is applied to highly dynamic surgical scenes with limited camera movement range, it is inclined to use floating or slim Gaussians to fit high-frequency image details.
Unfortunately, this overfitted representation suffers severe artifacts during simulation.
To tackle this problem, we present a novel geometrical regularization method in Section \ref{sec regular}, which leverages the stereo depth guidance and anisotropy regularization to obtain geometrically plausible scenes suitable for simulation.
Based on the improved GS-based surgical scene representation, we develop MPM-based simulation in Section \ref{sec simulation} for physically based soft tissue simulation.
Fig.~\ref{overview} depicts an overview of our proposed simulation method.

\subsection{3D Gaussians for Endoscopic Scene Representation}

\label{sec representation}
We represent the endoscopic scene as a group of 3D Gaussians.
A 3D Gaussian could be parameterized as $(\boldsymbol{\mu},\boldsymbol{\Sigma},\boldsymbol{c},\boldsymbol{\sigma})$,
which are corresponding to the position, covariance matrix, color and opacity respectively.
These features enable a Gaussian to effectively represent a small area of endoscopic tissue.
The 3D Gaussian is optimized using gradients backpropagated from the rendering loss.
To render the endoscopic scene composed of 3D Gaussians from a specific view, the covariance matrix of Gaussian can be projected at first:
$
    \hat{\mathbf{\Sigma}} = \mathbf{J}\mathbf{W}\mathbf{\Sigma} \mathbf{W}^T\mathbf{J}^T,
    \label{eq::projection}
$
where $\mathbf{W}$ and $\mathbf{J}$ are matrices related to view transformation and projection, $\hat{\mathbf{\Sigma}}$ is the projected covariance matrix which is used to determine whether a pixel is influenced by current Gaussian.
After projection, the value of each pixel is computed as a weighted sum of the color features:
\begin{equation}
    C(\mathbf{p}) = \sum_{i\in N}\boldsymbol{c_i} \alpha_i \prod_{j=1}^{i-1} (1-\alpha_j),
    \label{eq::rendering}
\end{equation}
where $N$ is the number of Gaussians influencing the shading of pixel $\mathbf{p}$ and $\alpha_i= \sigma_i \mathrm{exp}(-\frac{1}{2}(\mathbf{p-\bm \mu_i})^T{\mathbf{\Sigma}^{\prime}}^{-1}_{i}(\mathbf{p-\bm \mu_i}))$.
To reconstruct an endoscopic scene without surgical tool occlusion,
we apply masked L1 loss function at pixel $\mathbf{p}$ utilizing the k-th ground truth image $\boldsymbol{I}_k$, the corresponding tool mask $\boldsymbol{M}_k$ and the rendering result $C(\mathbf{p})$:
\begin{equation}
    \mathcal{L}_{color}(\mathbf{p}) = (1-\boldsymbol{M}_{k}(\mathbf{p}))*|C(\mathbf{p})-\boldsymbol{I}_k(\mathbf{p})|.
\end{equation}
This loss function allows us to effectively utilize surgical video information, as areas occluded by instruments in one frame may be visible in other frames. 


\subsection{Geometrically Regularized Optimization for 3D Gaussian}
\label{sec regular}
In endoscopic scene, the limited movement range of camera makes it difficult for GS to learn the geometry information and then lead to overfitting.
To overcome this problem, we initialize the positions of Gaussians by reprojecting the depth maps $\{\boldsymbol{D}_{i}\}^{T}_{i=1}$ to get a reasonable geometric structure before the training.
During the optimization, we add the Huber loss between estimated depth maps and rendered depth maps to the objective function of original Gaussian Splatting framework.
The depth value of 3D Gaussian at pixel $\mathbf{p}$ shares the same computation pattern with Eq. \ref{eq::rendering}.
Our complete loss function is:
\begin{equation}
    \mathcal{L}(\mathbf{p})=\mathcal{L}_{color}(\mathbf{p})+\eta \mathcal{L}_{Huber}(\boldsymbol{D}_{k}(\mathbf{p}), \sum_{i\in N}d_i \alpha_i \prod_{j=1}^{i-1} (1-\alpha_j))
\end{equation}
where $d_i$ is the z value of the i-th Gaussian, $\boldsymbol{D}_{k}$ is the k-th estimated depth map and hyperparameter $\eta$ is used to control the strength of depth regularization.
The depth regularization enforces Gaussians to distribute near the tissue surface and penalize the generation of floater to reduce artifact.
The artifacts in simulation results can also be caused by the Gaussians with slender shape.
Therefore, we perform anisotropy regularization to prune Gaussians with very slim shapes during the optimization.
The scale ratio is used to determine whether a Gaussian should be pruned: $Ratio(p) = \mathrm{max}(\mathbf{S_p})/\mathrm{min}{(\mathbf{S_p})}$,
where $\mathbf{S_p}$ is the scaling tensor of Gaussian $p$. 
We prune Gaussians whose scale ratio surpasses the predefined threshold $\gamma$ during the training.

\subsection{Physically-based 3D Gaussians Simulation}
\label{sec simulation}
Before simulation, we perform the Gaussian padding for better simulation effect.
Directly simulating Gaussians without padding often leads to surface crash as shown in Fig. \ref{fig:ablation}.
We compute the opacity field $O$ using the following equation on a uniform $100\times100\times100$ Eulerian grid:
$O(\boldsymbol{x})=\sum_{i}\sigma_i \mathrm{exp}(-\frac{1}{2}(\boldsymbol{x-x_i})^T{\mathbf{\Sigma_i}}^{-1}(\boldsymbol{x-x_i}))$,
where $\boldsymbol{x}$ is the position of grid node and $\boldsymbol{x_i}$ is the center of Gaussian $i$ surrounding the grid node.
If the value of current node in opacity field is less than that of nodes closer to camera, it indicates that the current grid node is behind the tissue surface and we will pad a Gaussian at this place.

We integrate the Material Point Method into our framework to perform physically based tissue simulation on reconstructed scenes represented in 3D Gaussians.
The position of 3D Gaussians can be directly updated as lagrangian particles in MPM.
To let 3D Gaussians capture the deformation of material, \cite{xie2023physgaussian} propose to update the covariance matrix of Gaussian as follows: 
\begin{equation}
    \mathbf{{\Sigma}^{\prime}}=\mathbf{F}\mathbf{\Sigma}\mathbf{F}^{T},
\end{equation}
where $\mathbf{F}$ is the deformation gradient obtained from MPM solver, $\mathbf{\Sigma}$ is the initial covariance matrix and $\mathbf{{\Sigma}^{\prime}}$ is the updated covariance matrix.
We utilize Neo-Hookean\cite{rivlin1948} constitutive model in MPM to model the nonlinear stress-strain relationship of biological tissue.
The first Piola-Kirchoff stress of Neo-Hookean model, denoted as $\mathbf{PK_1}$, is computed as follows:
\begin{equation}
    \mathbf{PK_1}=\mu(\mathbf{F}-\mathbf{F^{T}})+\lambda \mathrm{log}(J)\mathbf{F^{-T}},
\end{equation}
where $\mu$ and $\lambda$ are lame parameters computed using Young's modulus $E$ and Poisson ratio $\nu$ with following equations:
\begin{align}
        \mu=\frac{\mathrm{E}}{2(1+\mathrm{\nu})}
        , \,\lambda=\frac{\mathrm{E}\mathrm{\nu}}{(1+\mathrm{\nu})(1-2\mathrm{\nu})}.
\end{align}
$\mathbf{PK_1}$ will participate in the following computation of grid forces and then influence the tissue dynamic.
By adjusting the values of Young's modulus $E$ and Poisson ratio $\nu$, we can control the physical behavior of the endoscopic tissue, such as its stiffness, in a physically interpretable way.

%% file: sections/experiments.tex
\section{Experiments}\label{sec:exp}

We extensively evaluated our proposed data-driven surgical simulation on five different endoscopic scenes,
using stereo endoscopic videos introduced in\cite{ye2017selfsupervised}, EndoNeRF datasets~\cite{wang2022neural} and our in-house data.
These videos encompassed common surgical procedures involving the manipulation, pushing and pulling of deformable tissues.
We compared our method with an GS-based baseline.
Similar to \cite{zhu2024endogs,liu2024endogaussian,huang2024endo4dgs}, we quantitatively evaluated the surgical scene reconstruction using PSNR.
For the simulation performance, since it is difficult for a quantitative evaluation, we reported the simulation efficiency and qualitatively measured the simulation performance by comparing the texture and geometry details under different simulation interactions.
We also conducted an ablation study to evaluate the proposed depth supervision module, anisotropy regularization module and Gaussian padding module through qualitative comparison.


\subsection{Implementation Details}

In our implementation, we set the delta in Huber loss to 0.2, the weight of depth loss $\eta=0.3$ and the anisotropy regularization $\gamma=10$.
All endoscopic scenes used in our experiments are optimized with 7000 iterations.
We utilize STTR-light\cite{Li_2021_ICCV} pretrained on Scene Flow to estimate stereo depth maps of binocular surgical video frames.
The tool masks are automatically generated by SegmentAnything \cite{kirillov2023segment}.
For simulation, we perform 80 substeps in each step and the timestep of each substep is 0.0005s.
All experiments were conducted on a computer equipped with an Intel(R) Xeon(R) W-2223 CPU and RTX3090 GPU.

\subsection{Qualitative and Quantitative Results}
\begin{figure}[t]
\centering
\includegraphics[width=1.0\textwidth]{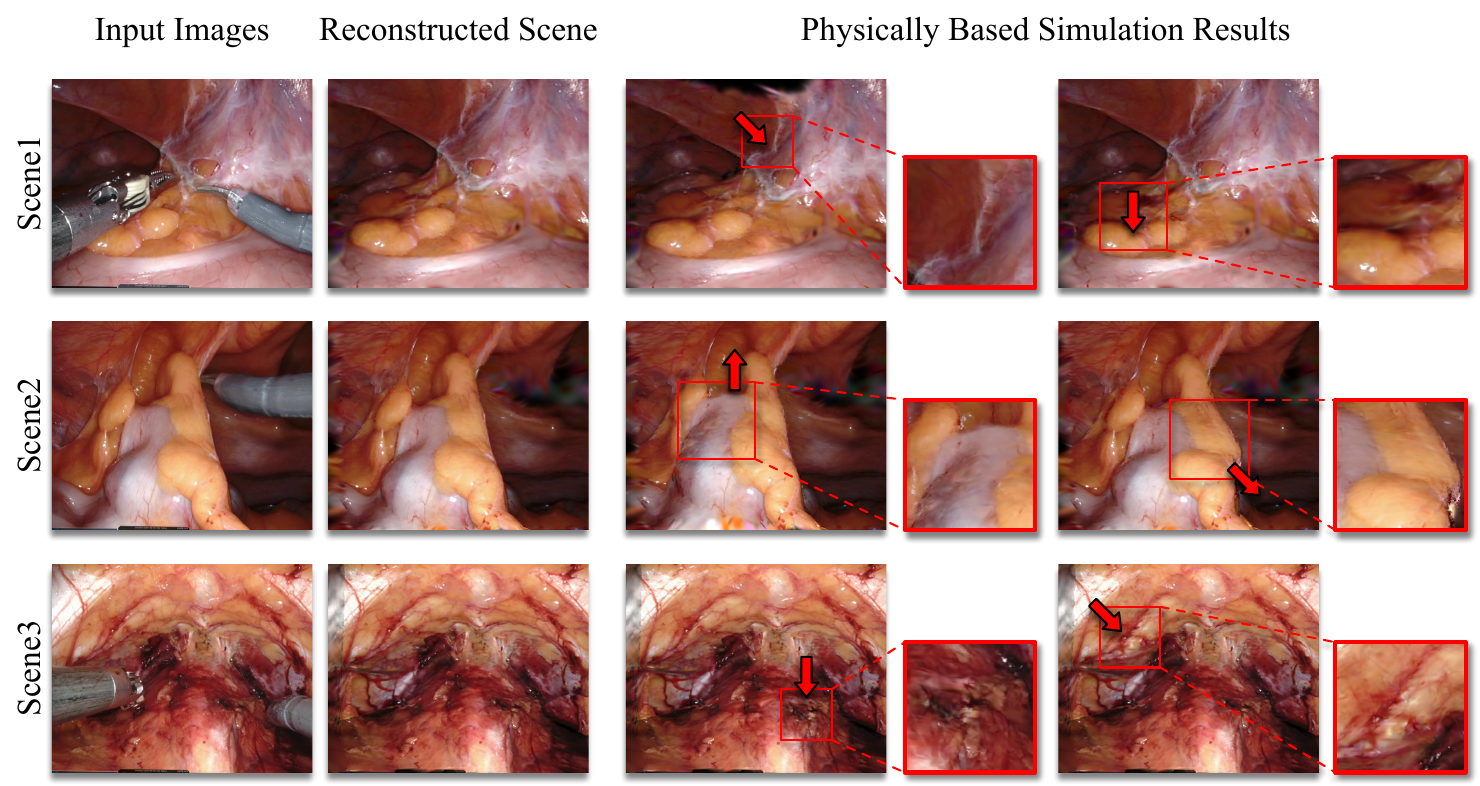}
\caption{{\bf Qualitative evaluation of simulation performance.} 
The external forces are indicated using red arrows.
The corresponding simulation videos are included in the supplementary material.}
\label{fig:results}
\end{figure}

We evaluated the proposed data-driven surgical scene simulation on various surgical scenarios.
Fig.~\ref{fig:results} presents the qualitative evaluation results.
Given the stereo surgical video, in which the tissue would be frequently occluded by the tool due to the surgical operation, 
our method can robustly reconstruct occlusion-free tissues with realistic textures.
It is the basis for achieving high-fidelity surgical simulation.
To assess the quality of the simulation, we simulated interactive actions with the reconstructed simulation environment by applying forces in various directions at different positions.
As shown in Fig.~\ref{fig:results}, a closer examination of the highlighted areas reveals that our simulation method not only generates visually plausible shape deformations but also preserves the visual consistency of the texture.
Table \ref{tbl:res} further reports the detailed processing time and simulation efficiency for each endoscopic scene.
Without heavy manual adjustments, our method is highly scalable and is able to efficiently consume various surgical videos for high-quality surgical simulation.
We refer readers to the supplementary video for more simulation results.

\begin{table*}[t]
  \centering
  \caption{\textbf{Quantitative results (standard
deviation in parentheses).}}
  \begin{tabularx}{\textwidth}{*{7}{>{\centering\arraybackslash}X}}
    \toprule
    Scenes & Training  &  Ours PSNR& EndoGS PSNR& Processing & Gaussian Number & Simulation FPS \\
    \midrule
    Scene1 & $65.0(0.8)$s & $34.77(0.04)$ & 40.44(0.30) &0.423(0.005)s & 78,712 & 20 \\
    Scene2 & $61.3(0.5)$s & $37.82(0.02)$ & 39.94(0.24) &0.495(0.006)s & 74,484 & 21 \\
    Scene3 & $60.7(0.5)$s & $36.05(0.24)$ & 36.28(0.04) &0.453(0.002)s & 58,223 & 25 \\
    \bottomrule
  \end{tabularx}
  \label{tbl:res}
\end{table*}

\begin{figure}[t]
\centering
\includegraphics[width=0.8\textwidth]{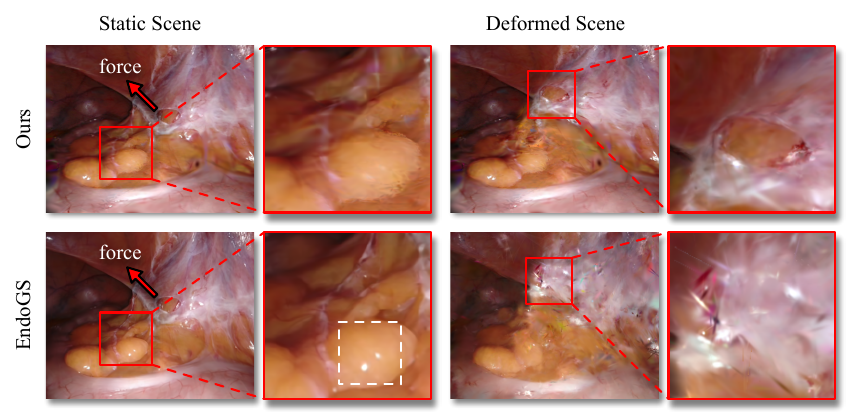}
\caption{\textbf{Comparison with EndoGS\cite{zhu2024endogs}.} The comparison between ours method and EndoGS on reconstruction and simulation.}
\label{fig:comparison}
\end{figure}

To further demonstrate the advantage of our method, we compared it with the EndoGS baseline\cite{zhu2024endogs}.
As shown in Table \ref{tbl:res}, we observed that EndoGS usually achieves a higher PSNR than ours, which indicates that its reconstructed surgical scene is closer to the original video frame.
We argue that it is because EndoGS is inclined to fit the high-frequency scene content~(e.g., specular light indicated by the white dashed box in Fig. \ref{fig:comparison}) with floater.
As shown in Fig. \ref{fig:comparison}, the reconstructed scene would result in obvious artifacts when it is applied to simulation.
In contrast, our method use geometric regularization to obtain scenes with more reasonable space structure.
It achieves comparable reconstruction quality to EndoGS.
More importantly, our method significantly outperforms EndoGS in terms of the simulation quality.
It not only maintains reasonable geometric structures and tissue textures but also produces physically realistic tissue deformations, as highlighted in the specific area of Fig. \ref{fig:comparison}.
These results fully demonstrate the superiority of our proposed method for endoscopic scene simulation.

\begin{figure}[t]
\centering
\includegraphics[width=0.8\textwidth]{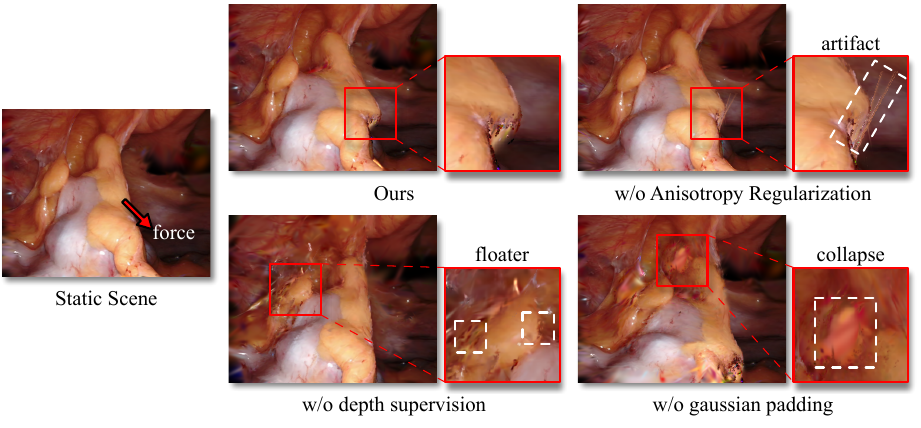}
\caption{\textbf{Ablation study.}  We compare our base simulation result and simulation results w/o depth supervision, gaussian padding or anisotropy regularization. The artifacts are highlighted using white dashed boxes.
}
\label{fig:ablation}
\end{figure}

To valid each module of our method, we conducted a qualitative ablation study.
Fig.~\ref{fig:ablation} presents the experimental results.
When anisotropy regularization is omitted, the slender kernels of the Gaussians become exposed under large-scale deformation, resulting in a fur-like artifact. 
In contrast to our results, simulations without depth supervision exhibit floating artifacts for lack of geometric regularization in training, leading to a noisy result after deformation. 
Simulations without Gaussian padding result in a thin surface that is prone to collapse under external forces. 
The corresponding result depicted in Fig. \ref{fig:ablation} exhibits noticeable cracks and exposes the underlying material.

